\pdfoutput=1

\documentclass[11pt]{article}
\usepackage{authblk}
\usepackage[]{EMNLP2023}

\usepackage{times}
\usepackage{latexsym}

\usepackage[T1]{fontenc}

\usepackage[utf8]{inputenc}

\usepackage{microtype}

\usepackage{inconsolata}

\usepackage{bibentry}
\usepackage{dialogue}
\usepackage{times}
\usepackage{latexsym}
\usepackage{mathrsfs}
\usepackage{amssymb}
\usepackage{amsmath}
\usepackage{url}
\usepackage{pifont}
\usepackage{graphicx}
\usepackage{color,xcolor}
\usepackage{braket}
\usepackage{bbold}
\usepackage[utf8]{inputenc}
\usepackage[toc,page]{appendix}
\usepackage{cleveref}
\usepackage{booktabs,graphicx,makecell,siunitx,eqparbox}
\usepackage[utf8]{inputenc}
\usepackage{color,soul}

\usepackage{enumitem}
\usepackage{float}

\title{A Digital Language Coherence Marker for Monitoring Dementia}

\author[1]{Dimitris Gkoumas}
\author[1,2]{Adam Tsakalidis}
\author[1,2]{Maria Liakata}
\affil[1]{Queen Mary University of London, London, UK}
\affil[2]{The Alan Turing Institute, London, UK}
\affil[ ]{\textit {\{d.gkoumas,a.tsakalidis,m.liakata\}@qmul.ac.uk}}

\begin{document}
\maketitle
\begin{abstract}
The use of spontaneous language to derive appropriate digital markers has become an emergent, promising and non-intrusive method to diagnose and monitor dementia. Here we propose methods to capture language coherence as a cost-effective, human-interpretable digital marker for monitoring cognitive changes in people with dementia. We introduce a novel task to learn the temporal logical consistency of utterances in short transcribed narratives and investigate a range of neural approaches. We compare such language coherence patterns between people with dementia and healthy controls and conduct a longitudinal evaluation against three clinical bio-markers to investigate the reliability of our proposed digital coherence marker. The coherence marker shows a significant difference between people with mild cognitive impairment, those with Alzheimer’s Disease and healthy controls. Moreover our analysis shows high association between the coherence marker and the clinical bio-markers as well as generalisability potential to other related conditions. 
\end{abstract}


\section{Introduction}
\label{intro}

Dementia includes a family of neurogenerative conditions that affect cognitive functions of adults. Early detection of cognitive decline could help manage underlying conditions and allow better quality of life. Many aspects of cognitive disorders manifest in the way speech is produced and in what is said \cite{forbes2005detecting,voleti2019review}. Previous studies showed that dementia is often associated with thought disorders relating to inability to produce and sustain coherent communication \cite{mckhann1987diagnostics,hoffman2020going}. Language coherence is a complex multifaceted concept which has been defined in different ways and to which several factors contribute \cite{redeker2000coherence}. A high-quality communication is logically consistent, topically coherent, and pragmatically reasonable \cite{wang2020narrative}.

\begin{figure}[hbt!]
\centering
\includegraphics[width=.46\textwidth]{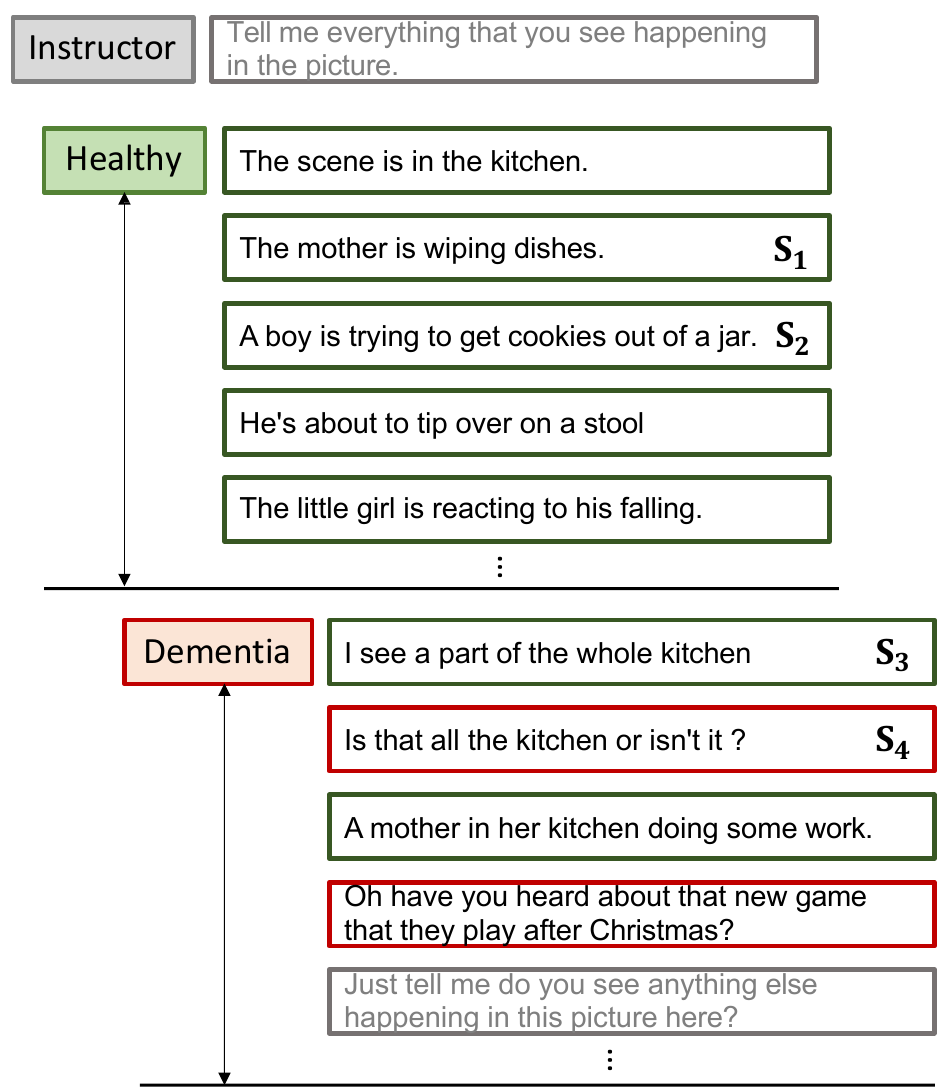}
\caption{Snapshots from healthy controls and people with dementia describing the Cookie Theft Picture. Green frames indicate logically consistent utterances and red disruptive ones (e.g., elaborations or `flight of ideas').}  
\label{fig:discource}
\end{figure}

Fig. \ref{fig:discource} illustrates two snapshots from people with dementia and healthy controls in the Pitt Corpus \cite{becker1994natural}, containing subjects' descriptions of the Cookie Theft Picture (CTP, Appx. \ref{app:ctp}) from the Boston Diagnostic Aphasia Examination \cite{ctp}.  As shown in Fig. \ref{fig:discource}, dementia subjects present more disruptions in the logical consistency of their CTP narratives than healthy controls. For example, 
the pair of semantically unrelated utterances $\{S_1,S_2\}$ is logically consistent and descriptive. By contrast, 
even though $\{S_3,S_4\}$ are semantically related, the pair is logically inconsistent since the latter utterance disrupts the description of the CTP. Here we focus on learning coherence as logical-thematic consistency of utterances in narratives, rather than the semantic relatedness of entities across sentences, to capture \textit{disruptive} utterances, such as  \textit{flight of ideas} and  \textit{discourse elaborations}. The latter have been shown to be indicative of cognitive disorders \cite{abdalla2018rhetorical,iter2018automatic}. Indeed, thought disorders (TD) is exhibited as disruption in the structure of thoughts and as it affects both language content and the thinking process, it affects how thoughts are expressed in language. TD is associated with various conditions including dementia. In particular, disorganized speech is a symptom of dementia and can be caused by damage to the brain that occurs with the disease~\cite{botha2019primary}.

The use of computational linguistics and natural language processing (NLP) to screen and monitor dementia progression has become an emergent and promising field \cite{fraser-meltzer-rudzicz:2016:JAD,koenig2018}.
However, recent work used language to distinguish people with Alzheimer’s Disease (AD) from  healthy controls, neglecting the longitudinal and fine-grained aspects of subjects' language impairments \cite{luz2020alzheimer,luz2021detecting,nasreen2021detecting}. Here, we address this limitation by first learning the  logical-thematic coherence of adjacent utterances in narratives, and then investigating the connection between longitudinal changes in language coherence and cognitive status.

Recent work for coherence in text has exploited deep \cite{cui2017text,feng2021towards}, discriminative \cite{xu2019cross}, and generative \cite{laban2021can} neural models for three evaluation tasks namely: a) the shuffle task (i.e., to discriminate genuine from randomly shuffled text), b) sentence ordering (i.e., to produce the correct order of sentences in a text) , and c) insertion (i.e., to predict the position of a missing sentence in a text). However these tasks are prone to learning the shuffle-ness of a text rather than its actual coherence \cite{laban2021can}. 
By contrast, our motivation is to learn
the logical consistency of adjacent utterances in narratives to capture fine-grained coherence impairments (Fig.~\ref{fig:discource}) rather than semantic relatedness or the global aspects of utterances’ order.
In this paper we make the following contributions:
\begin{itemize}[noitemsep,topsep=0pt,parsep=0pt,partopsep=0pt,leftmargin=*]
\item We define the new task of learning logical thematic coherence scores on the basis of the logical-thematic consistency of adjacent utterances (Sec.~\ref{sec:tasks}). We train on narratives from healthy controls in the DementiaBank Pitt Corpus \cite{becker1994natural}, hypothesising that controls produce a logically consistent order of utterances. We investigate a range of state-of-the-art (SOTA) neural approaches and obtain models in three different settings: a) fine-tuning transformer-based models, b) fully training discriminative models, and c) zero-shot learning with transformer-based generative models (Sec.~\ref{sec:coh_models}). Our experiments show that a fine-tuned transformer model (RoBERTa) achieves the highest discrimination between adjacent and non-adjacent utterances within a healthy cohort (Sec.~\ref{sec:quantitive}).

\item We introduce a human-interpretable digital coherence marker for dementia screening and monitoring from longitudinal language data. We first obtain logical thematic coherence scores of adjacent utterances and then aggregate these across the entire narrative (Sec.~\ref{sec:tasks}). 
\item We conduct a comprehensice longitudinal analysis to investigate how the digital coherence marker differs across healthy and dementia cohorts. 
The resulting digital coherence marker yields significant discrimination across healthy controls, people with mild cognitive impairment (MCI), and people with AD (Sec.~\ref{sec:discrimination}). 
\item We compare our digital coherence marker against one based on semantic similarity, showing superior performance of the former in both distinguishing across cohorts (Sec.~\ref{sec:discrimination}) and in detecting human-annotated disruptive utterances (Sec.~\ref{sec:human}). 
\item We evaluate our logical thematic coherence marker against three clinical bio-markers for cognitive impairment, showing high association and generalisability potential (Sec.~\ref{sec:biomarkers}). 

\end{itemize}

\section{Related Work}

\textbf{NLP and dementia:} Early NLP work for dementia detection analysed aspects of language such as lexical, grammatical, and semantic features \cite{ahmed2013connected,orimaye2017predicting,kave2018severity}, and studied para-linguistic features \cite{gayraud2011syntactic,lopez2013selection,pistono2019happens}. Recent work in this area has made use of manually engineered features \cite{luz2020alzheimer,luz2021detecting,nasreen2021detecting}, disfluency features \cite{nasreen2021alzheimer,rohanian2021alzheimer}, or acoustic embeddings \cite{yuan2020disfluencies,shor2020towards,pan2021using,zhu2021wavbert}. 
Closer to the current study, \citet{abdalla2018rhetorical} investigated discourse structure in people with AD by analyzing discourse relations. All such previous work has focused on differentiating across cohorts at fixed points in time without considering language changes over time.

\noindent \textbf{Coherence modeling:} The association between neuropsychological testing batteries and language leads researchers to exploit linguistic features and naive approaches for capturing coherence in spontaneous speech to predict the presence of a broad spectrum of cognitive and thought disorders. 
\cite{elvevaag2007quantifying,bedi2015automated,iter2018automatic}. Other work for coherence in text focused on feature engineering to implement some of the intuitions of Centering Theory \cite{lapata2005automatic,barzilay2008modeling,elsner2011extending,guinaudeau2013graph}. Despite their success, existing models either capture semantic relatedness or entity transition patterns across sentences rather than logical-thematic consistency. 
\noindent\textbf{Neural coherence:} Driven by the success of deep neural networks, researchers exploited distributed sentences \citet{cui2017text}, discriminative \citet{xu2019cross}, and BERT-based \citet{feng2021towards} models by evaluating coherence mostly on the shuffle task (refer to Sec. \ref{intro} for more details). Recent work has shown that a zero-shot setting in generative transformers can be more effective than fine-tuning BERT or RoBERTa achieving a new SOTA performance for document coherence \cite{laban2021can}. Here, we investigate a variety of such successful architectures to learn the temporal logical-thematic consistency of utterances in transcribed narratives. 
\section{Methodology}
\subsection{Logical Thematic Coherence}
\label{sec:tasks}
Let us denote a collection $C$ of $N$ transcribed narratives from healthy controls, i.e., $C= \{d_k\}_{k=1}^N$, where each narrative consists of a sequence of utterances $\{u_i\}$. The logical thematic coherence task consists in learning scores from adjacent pairs of utterances $(u_i,u_{i+1})$ in the healthy controls, so that these are higher than corresponding non-adjacent pairs of utterances $(u_i,u_{j})$ in a narrative, where $u_{j}$ is any forward utterance  following the adjacent pair \cite{feng2021towards} 

To monitor changes in cognition over time, we define a digital language coherence marker by computing the logical thematic coherence scores of adjacent utterances in people with dementia and controls in a test set and aggregating these over the entire narrative.  
To obtain comparisons across cohorts, we calculate longitudinal changes in the coherence marker from the last to the first and between adjacent subjects' narratives over the study. To assess the reliability of the coherence marker, we compute changes in the coherence marker and in widely used clinical markers from the end to the beginning of the study.

\subsection{Data}
\label{sub:data}
We have conducted experiments and trained coherence models on the DementiaBank Pitt Corpus \cite{becker1994natural}, where subjects are asked to describe the Cookie Theft picture \cite{ctp} up to 5 times across a longitudinal study (see Appx. \ref{app:pitt} for more details about the Pitt Corpus). \textbf{Coherent pairs:} We have learnt the temporal logical-thematic coherence of adjacent utterances from the healthy cohort, consisting of 99 people with a total amount of 243 narratives. \textbf{Incoherent pairs:} We use logically inconsistent utterance ordering by choosing utterances following an adjacent pair, from the same narrative so as to avoid learning cues unrelated to coherence due to potential differences in language style \cite{patil2020towards,feng2021towards}. 
While the level of coherence of controls may vary, we hypothesise that adjacent sentences by healthy controls will be more coherent than the negative instances, i.e. non-adjacent pairs from the same narrative. 
Table \ref{tbl:data_stats} summarizes the overall amount of utterances after splitting the healthy population into 80\%, 10\%, and 10\% for training, validation, and testing.  

\begin{table}[htbp]

\centering 

\resizebox{\columnwidth}{!}{%
\begin{tabular}{lccc}
 \toprule
 
\textbf{Utterances} & \textbf{Training} & \textbf{ Validation} & \textbf{Testing}    \\
 \hline
 \# Coherent &  2,178& 223 & 233 \\ \hline
 \# Incoherent & 16,181& 1,401 & 1,417\\

 \bottomrule
\end{tabular}
}
\caption{Amount of coherent and incoherent utterances for learning logical thematic coherence from the healthy cohort. } 
\label{tbl:data_stats}
\end{table}

To evaluate the ability of the digital language coherence marker to differentiate across cohorts and its reliability against the clinical bio-markers, we filtered people with dementia who have at least two narratives across the longitudinal study. This resulted in 62 people with AD and 14 people with MCI, with a total of 148 and 42 narratives respectively. We also included healthy controls, a total of 19 people with a total of 25 narratives.

\subsection{Coherence Models}
\label{sec:coh_models}

\noindent\textbf{Baseline Digital Marker:} We use Incoherence Model \cite{iter2018automatic}, which scores adjacent pairs of utterances in a narrative based on the cosine similarities of their sentence embeddings \cite{reimers-2019-sentence-bert}. 
We consider three main neural architectures, known to achieve SOTA performance on document coherence, to learn logical thematic coherence: A) fine-tuning transformer-based models, B) fully training discriminative models, and C) zero-shot learning with generative models. 

\paragraph{Transformer-based Models:} We fine-tune pre-trained transformers by maximising the probability that the second utterance in a pair follows the first (see Fig. \ref{fig:coherence_models} (A) in Appx. \ref{app:architectures}). The model's input is a sequence of tokens in the form of  $[CLS] + Utterance_1 + [SEP] + Utterance_2$,
where $(Utterance_1,Utterance_2)$ is a pair of either coherent of incoherent utterances in a narrative (see Sec.~\ref{sub:data}), $[SEP]$ is an utterance separator token, and $[CLS]$ is a pair-level token, used for computing the coherence score. We append to the transformer module a feed-forward neural network (FFNN) followed by a sigmoid function where the coherence score $f$ is the sigmoid function of FNNN that scales the output between 0 and 1. We fine-tune the  models with a standard binary cross-entropy loss function (i.e., BCELoss), setting the output of the model to 1 for coherent and 0 for incoherent pairs of utterances. 

We have experimented with the following variants: a) BERT-base \cite{lee2018pre} since it has been pre-trained on the Next Sentence Prediction (NSP) task which is similar to the task of scoring the coherence of adjacent utterances. b) RoBERTa-base \cite{liu2019roberta}, which has been pre-trained without the NSP task. c) a Convolutional Neural Network baseline \cite{cui2017text} which uses pre-trained word embeddings extracted by BERT-base (refer to Appx. \ref{app:architectures} for a detailed description).

\paragraph{Discriminative  Models:} We have trained discriminative models by maximizing the probability of an utterance pair being coherent. We have experimented with an architecture previously shown effective in coherence modelling for both speech \cite{patil2020towards} and text. \cite{xu2019cross}.

The model receives a pair of utterances and a sentence encoder maps the utterances to real-value vectors $U_1$ and $U_2$ (see Fig. \ref{fig:coherence_models} (B) in Appx. \ref{app:architectures}). The model then computes the concatenation of the two encoded utterances, as follows: 
\begin{equation}
   concat[U_1,U_2, U_1-U_2, U_1 * U_2, |U_1-U_2|]
   \label{eq:concat}
 \end{equation},
where $U_1-U_2$ is the element-wise difference, $U_1 * U_2$ is the element-wise product, and $|U_1-U_2|$ is the absolute value of the element-wise difference between the two encoded utterances. The choice to represent the difference between utterances in the form of Eq. \ref{eq:concat} was introduced  by \citet{xu2019cross} as a high level statistical function that could capture local level interaction between utterances and we make the same assumption. Finally, the concatenated feature representation is fed to a one-layer MLP to output the coherence score $f$. We have trained the model in bi-directional mode with inputs $(U_1,U_2)$ and $(U_2,U_1)$ for the forward and backward operations and used a margin loss as follows:
\begin{equation}
   L(f^+,f^-) = max(0, n - f^+ + f^-)
 \end{equation},
where $f^+$ is the coherence score of a coherent pair of utterances, $f^-$ thescore of an incoherent pair, and $n$ the margin hyperparameter.
The model can work with any pre-trained sentence encoder. Here, we experiment with two variants: a) pre-trained sentence embeddings from SentenceBERT \cite{reimers-2019-sentence-bert}(\textbf{DCM-sent}), and b) averaged pre-trained word embeddings extracted from BERT-base \cite{lee2018pre}(\textbf{DCM-word}).

\paragraph{Generative Models:}
We experiment with a zero-shot setting for generative transformers, an approach that previously achieved best out-of-the-box performance for document coherence \cite{laban2021can}. We provide a pair of utterances to a generative transformer and compute the perplexity in the sequence of words for each pair (refer to Appx. \ref{app:architectures} for a detailed description). Perplexity is defined as the exponential average log-likelihood in a sequence of words within a pair $P$ as follows:
\begin{equation}
\label{eq:perplexity}
   PPL(P) = exp \Bigl\{ -\frac{1}{t} \sum_{i}^t p(w_i|w_{<i})\Bigl\},
 \end{equation},
\noindent where $p(w_i|w_{<i})$ is the likelihood of the $i^{th}$ word given the preceding words $w_{<i}$ within a pair of utterances. Finally, we approximate the coherence score $f$ as follows: 
\begin{equation}
   f = 1- PPL(P),
\label{eq:per}   
 \end{equation}
We use $1-PPL$ rather than $PPL$ since low perplexity indicates that a pair is likely to occur, but we need high coherence scores for sequential pairs.

We have experimented with two SOTA generative transformers, of different sizes and architecture: a) \textbf{GPT2}, a decoder transformer-based model \cite{radford2019language} and b) \textbf{T5}, an encoder-decoder transformer-based model \cite{raffel2020exploring}. In the end we also pre-train T5-base, i.e., \textbf{T5-base$_{pre}$}. In particular, we feed sequential pairs of utterances and consider the loss on the second sequential sentence within the pair, just like sequence to sequence models. For testing, we extract coherence scores according to Eq. \ref{eq:per} for coherent and incoherent pairs. 

For the training details of coherence models please refer to Appx. \ref{app:train}.

\subsection{Evaluation Metrics}
\label{sec:metrics}
For evaluating the temporal logical thematic coherence models, we report the average coherence score of adjacent and non-adjacent utterance pairs, denoted as $f^+$ and $f^-$, respectively. The higher the $f$ score, the more coherent the pair. We also report the models' accuracy on adjacent utterances denoted as \textit{temporal} accuracy, i.e., $Acc_{temp}$, calculated as the correct rate between the adjacent utterances recognized as coherent and the total number of adjacent pairs in the test corpus. In particular, a pair of adjacent utterances $\{u_i, u_{i+1}\}$ in the test set is perceived as coherent if its coherence score $f_{(u_i,u_{i+1})}$ is higher than the coherence score $f_{(u_i,u_{k>i+1})}$  of the corresponding non-adjacent pair of utterances as follows:

\begin{equation}
  f({u_i, u_{i+1}}) = \begin{cases}
    1 & \text{if $f_{(u_i,u_{i+1})}$ > $f_{(u_i,u_{k>i+1})}$} \\
    0 & \text{otherwise}
  \end{cases}
\end{equation}
,
where $1$ corresponds to coherent and $0$ to incoherent pair, correspondingly. The coherence across an entire narrative is approximated by averaging the coherence scores of adjacent utterances, denoted as \textit{entire} accuracy, i.e., $Acc_{entire}$. Similarly, the entire accuracy is calculated as the correct rate of narratives recognized as coherent out of the total amount of narratives in the test corpus. A narrative is perceived as coherent if the averaged scores of the adjacent utterances are higher than the average scores of the non-adjacent ones within a narrative. The higher the temporal and entire accuracy, the better the model. Finally, we report the absolute percentage difference in $f$ scores between adjacent and non-adjacent utterances, denoted $\% \Delta$ (refer to Appx. \ref{app:difference} for more details), and the averaged loss of the models. The higher and more significant the $\% \Delta$, the better the model, while the reverse holds for the averaged loss.

To investigate the reliability of the digital coherence marker, we evaluate against three different clinical bio-markers collected from people with dementia. These are the Mini-Mental State Examination (MMSE), the Clinical Dementia Rating (CDR) scale \cite{morris1997clinical}, and the Hamilton Depression Rating (HDR) scale \cite{williams1988structured}. The lower the MMSE score the more severe the cognitive impairment. The opposite is true of the other scores, where a higher CDR score denotes more severe cognitive impairment and higher HDR scores indicate more severe depression (for more details about the bio-markers please refer to Appx. \ref{app:biomarkers}).
\section{Experimental Results}

\subsection{Logical Thematic Coherence Models}

\subsubsection{Quantitative Analysis}
\label{sec:quantitive}
\begin{table*}[htbp]
\centering 
\resizebox{\textwidth}{!}{%
\begin{tabular}{cccccccc}
 \toprule
\textbf{Model} & \textbf{Setting} & \textbf{Avg. $f^+$} & \textbf{Avg. $f^-$} & \textbf{$\% \Delta$} & \textbf{Avg. $Acc_{temp}$} & \textbf{Avg. $Acc_{entire}$} &  \textbf{Avg. Loss} \\
 \hline
CNN & Training & 0.560   & 0.475 & 18.2$^\dagger$  & 73.4\%  & 92.0\%   & 0.636\\
\hline 
BERT-base & Fine-tuning & 0.630   & 0.422 &  49.1$^\dagger$  & 75.4\%  &  \textbf{100.0\%}  &0.575  \\
\hline
RoBERTa-base & Fine-tuning & 0.604  & 0.353 & \textbf{71.0$^\dagger$}  & \textbf{81.4\%}  & \textbf{100.0\%}  &  \textbf{0.554}\\
\hline \hline 
DCM-sent & Training & -0.034 & -1.975& \textbf{98.2}$^\dagger$ &  63.9\% & 76.0\% &3.64 \\
\hline
DCM-word & Training & 0.282  & -1.068 & \textbf{126.4}$^\dagger$  & 69.6\%  & 80.0\%  & 3.84 \\
\hline \hline 
GPT2-base & Zero Shot & -383.8  & -384.8 & 0.3  &  50.4\%   & 48.0\%   & - \\
\hline
GPT2-medium & Zero Shot & -313.0   & -318.5 & 1.7  & 48.9\%  & 48.0\%   & -\\
\hline
GPT2-large & Zero Shot & -290.1  & -298.8 & -2.9  & 50.0\%  & 60.0\% & -\\
\hline
T5-base & Zero Shot &  -0.668  & -0.751 & 11.0  & 64.8\%  & 64.0\%  & - \\
\hline
T5-large & Zero Shot & -3.674  & -3.996&  8.1 & 58.2\%   &  60.0\%   & -\\
\hline
T5-base$_{pre}$ & Pre-train & -0.224 & -0.208 & 7.3  &  46.1\%    & 40.0\%  & 0.376 \\
 \bottomrule
\end{tabular}}
\caption{Performance of logical thematic coherence models trained on healthy controls in three different settings; A) training, B) fine-tuning, and C) zero-shot. $f^+$ is the coherence score of adjacent utterances, $f^-$ the coherence score of non-adjacent ones, and $\% \Delta$ the absolute percentage difference between $f^+$ and $f^-$. $\dagger$ denotes significant difference between the two coherence scores. $Acc_{temp}$ and $Acc_{entire}$ measure  accuracy on adjacent utterances and entire narratives, respectively. Best performance is highlighted in bold. } 
\label{tbl:results_healty}
\end{table*}

Table \ref{tbl:results_healty} summarizes the performance of logical thematic coherence models trained on the healthy cohort. Overall, fine-tuned transformerssignificantly outperform discriminative and generative transformer models. All models score higher on consecutive utterance pairs than non-consecutive ones.
While the absolute percentage difference of coherence scores between sequential and non-sequential pairs of utterances is higher for the discriminative models, $\% \Delta$ has a higher significance for the transformer-based models. 

BERT and RoBERTa are the best performing models, achieving a significant high entire accuracy (100\%), meaning that the model is able to predict all the narratives in the healthy population as being coherent, in line with our hypothesis. RoBERTa yielded an increased logical thematic coherence accuracy of 81.4\% compared to 75.4\% for BERT. Despite the original BERT being trained with two objectives, one of which is Next Sentence Prediction (NSP), an indirect signal for the coherence of adjacent utterances, RoBERTa, trained without the NSP objective, outperformed BERT. Presumably, RoBERTa outperforms BERT since the former was trained on a much larger dataset and using a more effective training procedure. Moreover, the simple CNN baseline, while performing worse than BERT and RoBERTa still outperforms the discriminative and generative models, which shows the effectiveness of fine-tuning.

The discriminative models perform better when using pre-trained embeddings from BERT rather than pre-trained sentence embeddings. Our experiments show that discriminative models are outperformed by transformers when modelling thematic logical coherence in transcribed narratives. This is contrary to  earlier work \cite{xu2019cross,patil2020towards} where discriminative models outperformed early RNN based models, but we note that this work did not compare against transformers. 

Despite \citet{laban2021can} showing that a zero-shot setting in generative transformers can be more effective than fine-tuning BERT or RoBERTa, our experiments show that this setting has the worst performance. The results did not improve even when we pre-trained the T5 model on the Pitt corpus (see T5-base$_{pre}$ in Table \ref{tbl:results_healty}). We presume that large pre-trained language models may suffer from domain adaptation issues here and operate on too short a window to capture logical consistency in narratives. Future work could investigate fine-tuning or prompt-training generative transformers for this task.



\subsection{The digital Language Coherence Marker}
Here, we exploited the best-performing logical thematic coherence model, i.e., RoBERTa, to obtain a digital language coherence marker for subjects across different cohorts over the longitudinal study (refer to Sec. \ref{sec:tasks} for more details). We first present results regarding the longitudinal discrimination ability for this marker and then show its reliability by evaluating against three clinical bio-markers.


\subsubsection{Longitudinal Discrimination Ability}
\label{sec:discrimination}


We analyzed changes in the digital marker over time and across cohorts. First, we calculated the average of digital markers across the three cohorts. The column $Marker$ in Table \ref{tbl:discr_marker} summarizes the results. The averaged digital marker was higher in the healthy cohort than in MCI and AD cohorts. Similarly, the averaged marker in the MCI group was higher than that in the AD group. However, the difference was significant only between the healthy and AD cohorts ($p < 0.05$) \footnote{\label{test}We use a nonparametric test, namely the Mann-Whitney test, to measure if the distribution of a variable is different in two groups.}.

\begin{table*}[htbp]

\centering 

\begin{tabular}{llcccccc}
 \toprule
 \multicolumn{1}{c}{} &
 \multicolumn{3}{c}{\textbf{Our digital marker}} &
 \multicolumn{3}{c}{\textbf{Baseline digital marker}} \\
\cmidrule(r){2-4} \cmidrule(r){5-7}

\textbf{Cohort} & \textbf{Marker} & \textbf{$\Delta_{(end-start)}$} & \textbf{$\Delta_{(long)}$}  & \textbf{Marker} & \textbf{$\Delta_{(end-start)}$} & \textbf{$\Delta_{(long)}$}     \\
 \hline
 Healthy & \textbf{0.604 (0.08)} &\textbf{ 0.09 (0.07)} & \textbf{0.07 (0.05)} & 0.249 (0.05) & 0.02 (0.06) & 0.01(0.06)\\ \hline
 MCI & 0.597 (0.09) & \textbf{-0.05 (0.09)} & \textbf{-0.05 (0.07)} &0.262 (0.06) & -0.03 (0.07) & -0.03 (0.06)\\  \hline
 AD & \textbf{0.567 (0.10)}  & \textbf{-0.02 (0.16)}  & \textbf{-0.02 (0.11)} & 0.241 (0.07) & -0.01 (0.08) & -0.01 (0.06) \\ 
 \bottomrule
\end{tabular}
\caption{ Longitudinal discrimination ability between the proposed digital marker and a baseline based on semantic similarity. Marker: Average of coherence marker within a population. $\Delta_{(end-start)}$: Average change of the marker from the end to the beginning of the study. $\Delta_{(long)}$: Average change of the digital marker between adjacent narratives within subjects. Numbers in  $()$ refer to corresponding standard deviations. Numbers in bold denote significant difference between the health controls and dementia cohorts (see Sec. \ref{sec:discrimination}).}
\label{tbl:discr_marker}
\end{table*}

We subsequently calculated changes in the digital marker from the end to the start of the study and across the cohorts (i.e.,  $\Delta_{(end-onset)}$ in Table \ref{tbl:discr_marker}). There was a significant decrease for the MCI and AD groups and a significant increase for the healthy controls ($p < 0.05$) $^{\ref{test}}$. The increase in healthy controls is presumably because subjects are able to remember and do better at the CTP description when seeing it again \cite{goldberg2015practice}. Moreover, we noticed that people with MCI exhibited more substantial change than those with AD, despite the average digital coherence marker of the former being 0.597 compared to 0.567 for the latter.

We also calculated changes in the digital marker between adjacent narratives over time and then aggregated the changes within subjects in the study. In Table \ref{tbl:discr_marker}, we report the average change across cohorts, i.e., $\Delta_{(long)}$. We obtain similar results as the ones taken from end to start.

We finally compared the longitudinal discrimination ability of our proposed digital marker with a baseline digital marker based on the semantic relatedness of adjacent utterances (refer to Sec. \ref{sec:coh_models}). The averaged baseline marker was higher in the MCI cohort than in healthy and AD cohorts (see Table \ref{tbl:discr_marker}). Moreover, there was no significant difference across the cohorts. On the other hand, we observed similar changes (i.e., $\Delta_{(end-start)}$ and $\Delta_{(long)}$ in Table \ref{tbl:discr_marker}) in the baseline marker over time compared to the one proposed in this paper. However, such changes were not significant across cohorts for the baseline marker ($p > 0.05$) $^{\ref{test}}$.

\subsubsection{Evaluation on Human-Annotated Disruptive Utterances}
\label{sec:human}
We investigated the effectiveness of the digital coherence marker in capturing disruptive utterances in narratives, and compared it with the baseline digital marker. Such disruptive utterances are annotated with the code $[+ \ exc]$ in the transcripts of the Pitt corpus and constitute a significant indicator of AD speech \cite{abdalla2018rhetorical,voleti2019review}. Out of 1,621 pairs of adjacent utterances in the AD cohort, 543 ones (33\%) are disruptive. For the baseline marker, the average score of disruptive utterances decreased to 0.19 (STD=0.17) compared to 0.26 (STD=0.17) for non-disruptive ones, i.e., an absolute percentage difference \footnote{For the definition refer to \ref{sec:metrics}.} of 31\%. For our proposed
marker, the average score of disruptive utterances decreased to 0.41 (STD=0.09) from 0.64 (STD=0.15) for non-disruptive ones, i.e., an absolute percentage difference of 44\%. The results showed that both digital markers significantly  captured disruptive utterances ($p_{t-test}<0.05$). However, our proposed digital marker is more robust in capturing such utterances.

\subsubsection{Association with Clinical Bio-markers}
\label{sec:biomarkers}
We investigated the reliability of the digital marker by associating its changes with different degrees of changes in cognitive status from the end to the beginning of the longitudinal study, as expressed by widely accepted cognition scales. We analyzed association patterns in the largest cohort, i.e., the AD group consisting of 62 participants.

We first investigated the association between changes in the coherence marker against the Mini-Mental State Examination (MMSE) \cite{morris1997clinical}. MMSE ranges from 0-30. The higher the MMSE score, the higher the cognitive function (refer to Appx. \ref{app:biomarkers} for more details about MMSE). Here, we have split the AD population into four bins on the basis of the magnitude of MMSE change. Table \ref{tbl:ad_mmse} provides details regarding bin intervals and the association of changes between the MMSE and the digital coherence marker.

\begin{table}[htbp]
\centering 

\resizebox{\columnwidth}{!}{%
\begin{tabular}{lccc}
 \toprule
 
\textbf{Bin} & \# \textbf{Subjects} & \textbf{$\Delta$ MMSE} & \textbf{$\Delta$ Coherence}    \\
 \hline
 Low & 25 & [-6,2] &  -0.003 (0.089)\\ \hline
 Minor & 17 & [-12,-7] & -0.030 (0.094)  \\ \hline
 Moderate & 11 & [-18,-13] & -0.076 (0.095)   \\ \hline
 Severe & 9 & [-27,-19] & -0.200 (0.104)   \\

 \bottomrule
\end{tabular}
}
\caption{Association between changes in Mini-Mental State Examination (MMSE) and the digital coherence marker in AD patients at different degrees of cognitive decline. Numbers in $[,]$ define the lower and upper values of each  bin interval. Numbers in $()$ refer to the standard deviation. $\#$ Subjects = Population within bins. $\Delta$ = Change from the end to the onset of the study.   } 
\label{tbl:ad_mmse}
\end{table}

Overall, we observed that the digital marker decreases across the population for the different degrees of cognitive decline. In particular, the higher the difference in MMSE, the more substantial the decrease in the digital marker change over the longitudinal study. For people with moderate or severe cognitive decline, the coherence decreased significantly compared to that of people with low cognitive decline ($p < 0.05$ ) $^{\ref{test},}$\footnote{\label{distribution}Here, we investigated how coherence change distributions differ across the AD population at different degrees of cognitive decline progression.}.

Next, we investigated the association between changes in the coherence marker and the Clinical Dementia Rating (CDR) \cite{morris1997clinical}. CDR is based on a scale of 0–3 in assessing people with dementia. The higher the CDR, the lower the cognitive function (refer to Appx. \ref{app:biomarkers} for more details about CDR). Here, we split the AD population into low, minor, moderate and severe bins according to the magnitude of CDR change, i.e., $\Delta$ CDR in Table \ref{tbl:ad_cdr}. The higher the CDR change the more severe the cognitive decline over time.

\begin{table}[htbp]
\centering 

\resizebox{\columnwidth}{!}{%
\begin{tabular}{lccc}
 \toprule
 
\textbf{Bin} & \# \textbf{Subjects} & \textbf{$\Delta$ CDR} & \textbf{$\Delta$ Coherence}    \\
 \hline
 Low & 20 & [0, 0.5] & -0.009 (0.091)\\ \hline
 Minor & 16 & (0.5,1.5] & -0.011 (0.060)   \\ \hline
 Moderate & 15 & (1.5,2.5] & -0.060 (0.110)   \\ \hline
 Severe & 11 & (2.5,3] &  -0.125 (0.078) \\

 \bottomrule
\end{tabular}
}
\caption{Association between changes in Clinical Dementia Rating (CDR) and the digital coherence marker in AD patients at different degrees of cognitive decline. Numbers in $(,]$ define the lower and upper values of each  bin interval. Numbers in $()$ refer to the standard deviation. $\#$ Subjects = Population within bins. $\Delta$ = Change from the end to the onset of the study.   } 
\label{tbl:ad_cdr}
\end{table}

The digital coherence marker decreased across the population at different degrees
of CDR change. In particular, the higher the increase in CDR, the higher the decrease in the digital coherence marker over the longitudinal study. Changes in the digital coherence marker are similar for people with low and minor cognitive decline. However, there is significant decrease in coherence for the moderate and severe bins compared to the minor and mild ones $p < 0.05$ ) $^{\ref{test}, \ref{distribution}}$ .

Finally, we investigated the generalisability potential of our proposed coherence marker in association with the Hamilton Depression Rating (HDR) \cite{williams1988structured}. HDR can be a useful scale for assessing cognitively impaired patients who have difficulty with self-report instruments  and is one of the most widely used and accepted instruments for assessing depression. It is based on a 17-item scale. The higher the HDR, the more severe the level of depression (refer to Appx. \ref{app:biomarkers} for more details about HDR). We investigated associations between the last HDR record \footnote{We considered the last  HDR record instead of changes in HDR over time since there were missing HDR measurements in the study. } and changes in the digital coherence marker from the end to start of the study. 
\begin{table}[htbp]
\centering 

\resizebox{\columnwidth}{!}{%
\begin{tabular}{lccc}
 \toprule
 
\textbf{Bin} & \# \textbf{Subjects} & \textbf{HDR} & \textbf{$\Delta$ Coherence}    \\
 \hline
 No Depression & 17 & [0,7] & -0.02 (0.11)\\ \hline
 Mild &  18& [8,16] &   -0.01 (0.10)  \\ \hline
 Moderate & 14  & [17,23] & -0.21 (0.10)   \\ 
 
 \bottomrule
\end{tabular}
}
\caption{Association between the last Hamilton Depression Rating (HDR) record and changes in the digital coherence  for AD patients. Numbers in $[,]$ define the lower and upper values of each  bin interval. Numbers in $()$ refer to the standard deviation. $\#$ Subjects = Population within bins. $\Delta$ = Change from the end to the onset of the study. } 
\label{tbl:ad_hdr}
\end{table}
Table \ref{tbl:ad_hdr} summarizes the association between HDR and changes in the digital coherence marker. Changes in coherence were similar for people with no or mild depression. However, there was a significant decrease for people with moderate depression ($p < 0.05$ ) $^{\ref{test}, \ref{distribution}}$. This is in line with current studies showing that individuals experiencing difficulty constructing coherent narratives generally report low well-being and more depressive symptoms \cite{vanderveren2020influence}.

\section{Conclusion}

We have introduced a new task for modelling the logical-thematic temporal coherence of utterances in short transcribed narratives to capture disruptive turns indicative of cognitive disorders. To this end, we have investigated transformer-based, discriminative, and generative neural approaches. Our experiments show that a fine-tuned transformer model (RoBERTa) achieves the best performance in capturing the coherence of adjacent utterances in narratives from the healthy cohort. We aggregate temporal language coherence to create a human-interpretable digital language coherence marker for longitudinal monitoring of cognitive decline. 
Longitudinal analysis showed that the digital marker is able to distinguish people with mild cognitive impairment, those with Alzheimer’s Disease (AD) and healthy controls. A comparison with a baseline digital marker based on semantic similarity showed the superiority of our digital marker. Moreover, evaluation against three clinical bio-markers showed that language coherence can capture changes at different degrees of cognitive decline and achieves significant discrimination between people with moderate or severe cognitive decline within an AD population. It can also capture levels of depression, showing generalisability potential. In future, we aim to integrate disfluency language patterns and develop strategies for improving the performance of generative models.


\section*{Limitations}

Monitoring dementia using computational linguistics approaches is an important topic. Previous work has mostly focused on distinguishing people with AD from healthy controls rather than monitoring changes in cognitive status per individual over time. In this study, we have used the Pitt corpus, currently the largest available longitudinal dementia dataset, to investigate longitudinal changes in logical coherence and their association with participants’ cognitive decline over time. An important limitation of the Pitt corpus is that the longitudinal aspect is limited, spanning up to 5 sessions/narratives maximum per individual with most participants contributing up to two narratives. Moreover, the number of participants is relatively small, especially for the MCI cohort. In the future, we aim to address these limitations by investigating the generalisability of the proposed digital language coherence marker on a recently introduced rich longitudinal dataset for dementia (currently under review) and on transcribed psychotherapy sessions (data is collected in Hebrew) to monitor mood disorders.

In this study, we used manually transcribed data from Pitt. In a real-world scenario, participants mostly provide speech via a speech elicitation task. This implies that the introduced method requires an automatic speech recognition (ASR) system robust to various sources of noise to be operationalized. ASR for mental health is currently underexplored, with most transcription work being done by human transcription.

It may be that the proposed digital coherence marker becomes a less accurate means for monitoring dementia when people experience other comorbidities, neurodegenerative and mental illnesses, that significantly affect speech and language. Indeed, cognitive-linguistic function is a strong biomarker for neuropsychological health \cite{voleti2019review}.

Finally, there is a great deal of variability to be expected in speech and language data affecting the sensitivity of the proposed digital marker. Both speech and language are impacted by speaker identity, context, background noise, spoken language etc. Moreover, people may vary in their use of language due to various social contexts and conditions, a.k.a., style-shifting \cite{coupland2007style}. Both inter and intra-speaker variability in language could affect the sensitivity of the proposed digital marker. While it is possible to tackle intra-speaker language variability, e.g., by integrating speaker-dependent information to the language, the inter-speaker variability remains an open-challenging research question.
\section*{Ethics Statement}

Our work does not involve ethical considerations around the analysis of the DementiaBank Pitt corpus as it is widely used. Ethics was obtained by the original research team by James Backer and participating individuals consented to share their data in accordance with a larger protocol administered by the Alzheimer and Related Dementias Study at the University of Pittsburgh School of Medicine \cite{becker1994natural}. Access to the data is password protected and restricted to those signing an agreement. 

This work uses transcribed dementia data to identify changes in cognitive status considering individuals’ language . Potential risks from the application of our work in being able to identify cognitive decline in individuals are akin to those who misuse personal information for their own profit without considering the impact and the social consequences in the broader community. Potential mitigation strategies include running the software on authorised servers, with encrypted data during transfer, and anonymization of data prior to analysis. Another possibility would be to perform on-device  processing (e.g. on individuals’ computers or other devices) for identifying changes in cognition and the results of the analysis would only be shared with authorised individuals. Individuals would be consented before any of our software would be run on their data.

\section*{Acknowledgements}
This work was supported by a UKRI/EPSRC
Turing AI Fellowship to Maria Liakata (grant
EP/V030302/1), the Alan Turing Institute (grant
EP/N510129/1), and Wellcome Trust MEDEA (grant 213939). Matthew Purver acknowledges financial support from the UK EPSRC via the projects Sodestream (EP/S033564/1) and ARCIDUCA (EP/W001632/1), and from the Slovenian Research Agency grant for research core funding P2-0103.

\bibliography{acl2023}

\begin{thebibliography}{49}
\expandafter\ifx\csname natexlab\endcsname\relax\def\natexlab#1{#1}\fi

\bibitem[{Abdalla et~al.(2018)Abdalla, Rudzicz, and Hirst}]{abdalla2018rhetorical}
Mohamed Abdalla, Frank Rudzicz, and Graeme Hirst. 2018.
\newblock Rhetorical structure and alzheimer’s disease.
\newblock \emph{Aphasiology}, 32(1):41--60.

\bibitem[{Ahmed et~al.(2013)Ahmed, Haigh, de~Jager, and Garrard}]{ahmed2013connected}
Samrah Ahmed, Anne-Marie~F Haigh, Celeste~A de~Jager, and Peter Garrard. 2013.
\newblock Connected speech as a marker of disease progression in autopsy-proven alzheimer’s disease.
\newblock \emph{Brain}, 136(12):3727--3737.

\bibitem[{Barzilay and Lapata(2008)}]{barzilay2008modeling}
Regina Barzilay and Mirella Lapata. 2008.
\newblock Modeling local coherence: An entity-based approach.
\newblock \emph{Computational Linguistics}, 34(1):1--34.

\bibitem[{Becker et~al.(1994)Becker, Boiler, Lopez, Saxton, and McGonigle}]{becker1994natural}
James~T Becker, Fran{\c{c}}ois Boiler, Oscar~L Lopez, Judith Saxton, and Karen~L McGonigle. 1994.
\newblock The natural history of alzheimer's disease: description of study cohort and accuracy of diagnosis.
\newblock \emph{Archives of neurology}, 51(6):585--594.

\bibitem[{Bedi et~al.(2015)Bedi, Carrillo, Cecchi, Slezak, Sigman, Mota, Ribeiro, Javitt, Copelli, and Corcoran}]{bedi2015automated}
Gillinder Bedi, Facundo Carrillo, Guillermo~A Cecchi, Diego~Fern{\'a}ndez Slezak, Mariano Sigman, Nat{\'a}lia~B Mota, Sidarta Ribeiro, Daniel~C Javitt, Mauro Copelli, and Cheryl~M Corcoran. 2015.
\newblock Automated analysis of free speech predicts psychosis onset in high-risk youths.
\newblock \emph{npj Schizophrenia}, 1(1):1--7.

\bibitem[{Botha and Josephs(2019)}]{botha2019primary}
Hugo Botha and Keith~A Josephs. 2019.
\newblock Primary progressive aphasias and apraxia of speech.
\newblock \emph{Continuum: Lifelong Learning in Neurology}, 25(1):101.

\bibitem[{Coupland(2007)}]{coupland2007style}
Nikolas Coupland. 2007.
\newblock \emph{Style: Language variation and identity}.
\newblock Cambridge University Press.

\bibitem[{Cui et~al.(2017)Cui, Li, Zhang, and Zhang}]{cui2017text}
Baiyun Cui, Yingming Li, Yaqing Zhang, and Zhongfei Zhang. 2017.
\newblock Text coherence analysis based on deep neural network.
\newblock In \emph{Proceedings of the 2017 ACM on Conference on Information and Knowledge Management}, pages 2027--2030.

\bibitem[{Elsner and Charniak(2011)}]{elsner2011extending}
Micha Elsner and Eugene Charniak. 2011.
\newblock Extending the entity grid with entity-specific features.
\newblock In \emph{Proceedings of the 49th Annual Meeting of the Association for Computational Linguistics: Human Language Technologies}, pages 125--129.

\bibitem[{Elvev{\aa}g et~al.(2007)Elvev{\aa}g, Foltz, Weinberger, and Goldberg}]{elvevaag2007quantifying}
Brita Elvev{\aa}g, Peter~W Foltz, Daniel~R Weinberger, and Terry~E Goldberg. 2007.
\newblock Quantifying incoherence in speech: an automated methodology and novel application to schizophrenia.
\newblock \emph{Schizophrenia research}, 93(1-3):304--316.

\bibitem[{Feng and Mostow(2021)}]{feng2021towards}
Jingrong Feng and Jack Mostow. 2021.
\newblock Towards difficulty controllable selection of next-sentence prediction questions.
\newblock In \emph{EDM}.

\bibitem[{Forbes-McKay and Venneri(2005)}]{forbes2005detecting}
Katrina~E Forbes-McKay and Annalena Venneri. 2005.
\newblock Detecting subtle spontaneous language decline in early alzheimer’s disease with a picture description task.
\newblock \emph{Neurological sciences}, 26(4):243--254.

\bibitem[{Fraser et~al.(2016)Fraser, Meltzer, and Rudzicz}]{fraser-meltzer-rudzicz:2016:JAD}
Kathleen~C. Fraser, Jed~A. Meltzer, and Frank Rudzicz. 2016.
\newblock Linguistic features identify {A}lzheimer's disease in narrative speech.
\newblock \emph{Journal of Alzheimer's Disease}, 49(2):407--422.

\bibitem[{Gayraud et~al.(2011)Gayraud, Lee, and Barkat-Defradas}]{gayraud2011syntactic}
Frederique Gayraud, Hye-Ran Lee, and Melissa Barkat-Defradas. 2011.
\newblock Syntactic and lexical context of pauses and hesitations in the discourse of alzheimer patients and healthy elderly subjects.
\newblock \emph{Clinical linguistics \& phonetics}, 25(3):198--209.

\bibitem[{Goldberg et~al.(2015)Goldberg, Harvey, Wesnes, Snyder, and Schneider}]{goldberg2015practice}
Terry~E Goldberg, Philip~D Harvey, Keith~A Wesnes, Peter~J Snyder, and Lon~S Schneider. 2015.
\newblock Practice effects due to serial cognitive assessment: implications for preclinical alzheimer's disease randomized controlled trials.
\newblock \emph{Alzheimer's \& Dementia: Diagnosis, Assessment \& Disease Monitoring}, 1(1):103--111.

\bibitem[{Goodglass et~al.(2001)Goodglass, Kaplan, Weintraub, and Barresi}]{ctp}
H.~Goodglass, E.~Kaplan, S.~Weintraub, and B.~Barresi. 2001.
\newblock The boston diagnostic aphasia examination.
\newblock \emph{Philadelphia, PA: Lippincott, Williams \& Wilkins}.

\bibitem[{Guinaudeau and Strube(2013)}]{guinaudeau2013graph}
Camille Guinaudeau and Michael Strube. 2013.
\newblock Graph-based local coherence modeling.
\newblock In \emph{Proceedings of the 51st Annual Meeting of the Association for Computational Linguistics (Volume 1: Long Papers)}, pages 93--103.

\bibitem[{Hoffman et~al.(2020)Hoffman, Cogdell-Brooke, and Thompson}]{hoffman2020going}
Paul Hoffman, Lucy Cogdell-Brooke, and Hannah~E Thompson. 2020.
\newblock Going off the rails: Impaired coherence in the speech of patients with semantic control deficits.
\newblock \emph{Neuropsychologia}, 146:107516.

\bibitem[{Iter et~al.(2018)Iter, Yoon, and Jurafsky}]{iter2018automatic}
Dan Iter, Jong Yoon, and Dan Jurafsky. 2018.
\newblock Automatic detection of incoherent speech for diagnosing schizophrenia.
\newblock In \emph{Proceedings of the Fifth Workshop on Computational Linguistics and Clinical Psychology: From Keyboard to Clinic}, pages 136--146.

\bibitem[{Kav{\'e} and Dassa(2018)}]{kave2018severity}
Gitit Kav{\'e} and Ayelet Dassa. 2018.
\newblock Severity of alzheimer’s disease and language features in picture descriptions.
\newblock \emph{Aphasiology}, 32(1):27--40.

\bibitem[{K{\"o}nig et~al.(2018)K{\"o}nig, Linz, Tr{\"o}ger, Wolters, Alexandersson, and Robert}]{koenig2018}
Alexandra K{\"o}nig, Nicklas Linz, Johannes Tr{\"o}ger, Maria Wolters, Jan Alexandersson, and Philippe Robert. 2018.
\newblock \href {https://doi.org/10.1159/000487852} {Fully automatic speech-based analysis of the semantic verbal fluency task}.
\newblock \emph{Dementia and Geriatric Cognitive Disorders}, 45(3-4):198--209.

\bibitem[{Laban et~al.(2021)Laban, Dai, Bandarkar, and Hearst}]{laban2021can}
Philippe Laban, Luke Dai, Lucas Bandarkar, and Marti~A Hearst. 2021.
\newblock Can transformer models measure coherence in text? re-thinking the shuffle test.
\newblock \emph{arXiv preprint arXiv:2107.03448}.

\bibitem[{Lapata et~al.(2005)Lapata, Barzilay et~al.}]{lapata2005automatic}
Mirella Lapata, Regina Barzilay, et~al. 2005.
\newblock Automatic evaluation of text coherence: Models and representations.
\newblock In \emph{IJCAI}, volume~5, pages 1085--1090. Citeseer.

\bibitem[{Lee and Toutanova(2018)}]{lee2018pre}
J~Devlin M Chang~K Lee and K~Toutanova. 2018.
\newblock Pre-training of deep bidirectional transformers for language understanding.
\newblock \emph{arXiv preprint arXiv:1810.04805}.

\bibitem[{Liu et~al.(2019)Liu, Ott, Goyal, Du, Joshi, Chen, Levy, Lewis, Zettlemoyer, and Stoyanov}]{liu2019roberta}
Yinhan Liu, Myle Ott, Naman Goyal, Jingfei Du, Mandar Joshi, Danqi Chen, Omer Levy, Mike Lewis, Luke Zettlemoyer, and Veselin Stoyanov. 2019.
\newblock Roberta: A robustly optimized bert pretraining approach.
\newblock \emph{arXiv preprint arXiv:1907.11692}.

\bibitem[{L{\'o}pez-de Ipi{\~n}a et~al.(2013)L{\'o}pez-de Ipi{\~n}a, Alonso, Travieso, Sol{\'e}-Casals, Egiraun, Faundez-Zanuy, Ezeiza, Barroso, Ecay-Torres, Martinez-Lage et~al.}]{lopez2013selection}
Karmele L{\'o}pez-de Ipi{\~n}a, Jesus-Bernardino Alonso, Carlos~Manuel Travieso, Jordi Sol{\'e}-Casals, Harkaitz Egiraun, Marcos Faundez-Zanuy, Aitzol Ezeiza, Nora Barroso, Miriam Ecay-Torres, Pablo Martinez-Lage, et~al. 2013.
\newblock On the selection of non-invasive methods based on speech analysis oriented to automatic alzheimer disease diagnosis.
\newblock \emph{Sensors}, 13(5):6730--6745.

\bibitem[{Luz et~al.(2020)Luz, Haider, de~la Fuente, Fromm, and MacWhinney}]{luz2020alzheimer}
Saturnino Luz, Fasih Haider, Sofia de~la Fuente, Davida Fromm, and Brian MacWhinney. 2020.
\newblock Alzheimer's dementia recognition through spontaneous speech: the adress challenge.
\newblock \emph{arXiv preprint arXiv:2004.06833}.

\bibitem[{Luz et~al.(2021)Luz, Haider, de~la Fuente, Fromm, and MacWhinney}]{luz2021detecting}
Saturnino Luz, Fasih Haider, Sofia de~la Fuente, Davida Fromm, and Brian MacWhinney. 2021.
\newblock Detecting cognitive decline using speech only: The adresso challenge.
\newblock \emph{arXiv preprint arXiv:2104.09356}.

\bibitem[{McKhann(1987)}]{mckhann1987diagnostics}
G~McKhann. 1987.
\newblock Diagnostics and statistical manual of mental disorders.
\newblock \emph{Arlington, VA: American Psychiatric Association}.

\bibitem[{Morris(1997)}]{morris1997clinical}
John~C Morris. 1997.
\newblock Clinical dementia rating: a reliable and valid diagnostic and staging measure for dementia of the alzheimer type.
\newblock \emph{International psychogeriatrics}, 9(S1):173--176.

\bibitem[{Nasreen et~al.(2021{\natexlab{a}})Nasreen, Hough, Purver et~al.}]{nasreen2021detecting}
Shamila Nasreen, Julian Hough, Matthew Purver, et~al. 2021{\natexlab{a}}.
\newblock Detecting alzheimer's disease using interactional and acoustic features from spontaneous speech.
\newblock Interspeech.

\bibitem[{Nasreen et~al.(2021{\natexlab{b}})Nasreen, Rohanian, Hough, and Purver}]{nasreen2021alzheimer}
Shamila Nasreen, Morteza Rohanian, Julian Hough, and Matthew Purver. 2021{\natexlab{b}}.
\newblock Alzheimer’s dementia recognition from spontaneous speech using disfluency and interactional features.
\newblock \emph{Frontiers in Computer Science}, page~49.

\bibitem[{Orimaye et~al.(2017)Orimaye, Wong, Golden, Wong, and Soyiri}]{orimaye2017predicting}
Sylvester~O Orimaye, Jojo~SM Wong, Karen~J Golden, Chee~P Wong, and Ireneous~N Soyiri. 2017.
\newblock Predicting probable alzheimer’s disease using linguistic deficits and biomarkers.
\newblock \emph{BMC bioinformatics}, 18(1):1--13.

\bibitem[{Pan et~al.(2021)Pan, Mirheidari, Harris, Thompson, Jones, Snowden, Blackburn, and Christensen}]{pan2021using}
Yilin Pan, Bahman Mirheidari, Jennifer~M Harris, Jennifer~C Thompson, Matthew Jones, Julie~S Snowden, Daniel Blackburn, and Heidi Christensen. 2021.
\newblock Using the outputs of different automatic speech recognition paradigms for acoustic-and bert-based alzheimer's dementia detection through spontaneous speech.
\newblock In \emph{Interspeech}, pages 3810--3814.

\bibitem[{Patil et~al.(2020)Patil, Singla, Shah, Hama, and Zimmermann}]{patil2020towards}
Rajaswa Patil, Yaman~Kumar Singla, Rajiv~Ratn Shah, Mika Hama, and Roger Zimmermann. 2020.
\newblock Towards modelling coherence in spoken discourse.
\newblock \emph{arXiv preprint arXiv:2101.00056}.

\bibitem[{Pistono et~al.(2019)Pistono, Pariente, B{\'e}zy, Lemesle, Le~Men, and Jucla}]{pistono2019happens}
Aur{\'e}lie Pistono, Jeremie Pariente, C~B{\'e}zy, B~Lemesle, J~Le~Men, and M{\'e}lanie Jucla. 2019.
\newblock What happens when nothing happens? an investigation of pauses as a compensatory mechanism in early alzheimer's disease.
\newblock \emph{Neuropsychologia}, 124:133--143.

\bibitem[{Radford et~al.(2019)Radford, Wu, Child, Luan, Amodei, Sutskever et~al.}]{radford2019language}
Alec Radford, Jeffrey Wu, Rewon Child, David Luan, Dario Amodei, Ilya Sutskever, et~al. 2019.
\newblock Language models are unsupervised multitask learners.
\newblock \emph{OpenAI blog}, 1(8):9.

\bibitem[{Raffel et~al.(2020)Raffel, Shazeer, Roberts, Lee, Narang, Matena, Zhou, Li, Liu et~al.}]{raffel2020exploring}
Colin Raffel, Noam Shazeer, Adam Roberts, Katherine Lee, Sharan Narang, Michael Matena, Yanqi Zhou, Wei Li, Peter~J Liu, et~al. 2020.
\newblock Exploring the limits of transfer learning with a unified text-to-text transformer.
\newblock \emph{J. Mach. Learn. Res.}, 21(140):1--67.

\bibitem[{Redeker(2000)}]{redeker2000coherence}
Gisela Redeker. 2000.
\newblock Coherence and structure in text and discourse.
\newblock \emph{Abduction, belief and context in dialogue}, 233(263).

\bibitem[{Reimers and Gurevych(2019)}]{reimers-2019-sentence-bert}
Nils Reimers and Iryna Gurevych. 2019.
\newblock \href {http://arxiv.org/abs/1908.10084} {Sentence-bert: Sentence embeddings using siamese bert-networks}.
\newblock In \emph{Proceedings of the 2019 Conference on Empirical Methods in Natural Language Processing}. Association for Computational Linguistics.

\bibitem[{Rohanian et~al.(2021)Rohanian, Hough, and Purver}]{rohanian2021alzheimer}
Morteza Rohanian, Julian Hough, and Matthew Purver. 2021.
\newblock Alzheimer's dementia recognition using acoustic, lexical, disfluency and speech pause features robust to noisy inputs.
\newblock \emph{arXiv preprint arXiv:2106.15684}.

\bibitem[{Shor et~al.(2020)Shor, Jansen, Maor, Lang, Tuval, Quitry, Tagliasacchi, Shavitt, Emanuel, and Haviv}]{shor2020towards}
Joel Shor, Aren Jansen, Ronnie Maor, Oran Lang, Omry Tuval, Felix de~Chaumont Quitry, Marco Tagliasacchi, Ira Shavitt, Dotan Emanuel, and Yinnon Haviv. 2020.
\newblock Towards learning a universal non-semantic representation of speech.
\newblock \emph{arXiv preprint arXiv:2002.12764}.

\bibitem[{Vanderveren et~al.(2020)Vanderveren, Aerts, Rousseaux, Bijttebier, and Hermans}]{vanderveren2020influence}
Elien Vanderveren, Loes Aerts, Sofie Rousseaux, Patricia Bijttebier, and Dirk Hermans. 2020.
\newblock The influence of an induced negative emotional state on autobiographical memory coherence.
\newblock \emph{Plos one}, 15(5):e0232495.

\bibitem[{Voleti et~al.(2019)Voleti, Liss, and Berisha}]{voleti2019review}
Rohit Voleti, Julie~M Liss, and Visar Berisha. 2019.
\newblock A review of automated speech and language features for assessment of cognitive and thought disorders.
\newblock \emph{IEEE journal of selected topics in signal processing}, 14(2):282--298.

\bibitem[{Wang et~al.(2020)Wang, Durrett, and Erk}]{wang2020narrative}
Su~Wang, Greg Durrett, and Katrin Erk. 2020.
\newblock Narrative interpolation for generating and understanding stories.
\newblock \emph{arXiv preprint arXiv:2008.07466}.

\bibitem[{Williams(1988)}]{williams1988structured}
Janet~BW Williams. 1988.
\newblock A structured interview guide for the hamilton depression rating scale.
\newblock \emph{Archives of general psychiatry}, 45(8):742--747.

\bibitem[{Xu et~al.(2019)Xu, Saghir, Kang, Long, Bose, Cao, and Cheung}]{xu2019cross}
Peng Xu, Hamidreza Saghir, Jin~Sung Kang, Teng Long, Avishek~Joey Bose, Yanshuai Cao, and Jackie Chi~Kit Cheung. 2019.
\newblock A cross-domain transferable neural coherence model.
\newblock \emph{arXiv preprint arXiv:1905.11912}.

\bibitem[{Yuan et~al.(2020)Yuan, Bian, Cai, Huang, Ye, and Church}]{yuan2020disfluencies}
Jiahong Yuan, Yuchen Bian, Xingyu Cai, Jiaji Huang, Zheng Ye, and Kenneth Church. 2020.
\newblock Disfluencies and fine-tuning pre-trained language models for detection of alzheimer's disease.
\newblock In \emph{INTERSPEECH}, volume 2020, pages 2162--6.

\bibitem[{Zhu et~al.(2021)Zhu, Obyat, Liang, Batsis, and Roth}]{zhu2021wavbert}
Youxiang Zhu, Abdelrahman Obyat, Xiaohui Liang, John~A Batsis, and Robert~M Roth. 2021.
\newblock Wavbert: Exploiting semantic and non-semantic speech using wav2vec and bert for dementia detection.
\newblock In \emph{Interspeech}, pages 3790--3794.

\end{thebibliography}
\bibliographystyle{acl_natbib}

\appendix
\section{The Cookie Theft Picture}
\label{app:ctp}

\begin{figure}[ht]
\centering
\includegraphics[width=.45\textwidth]{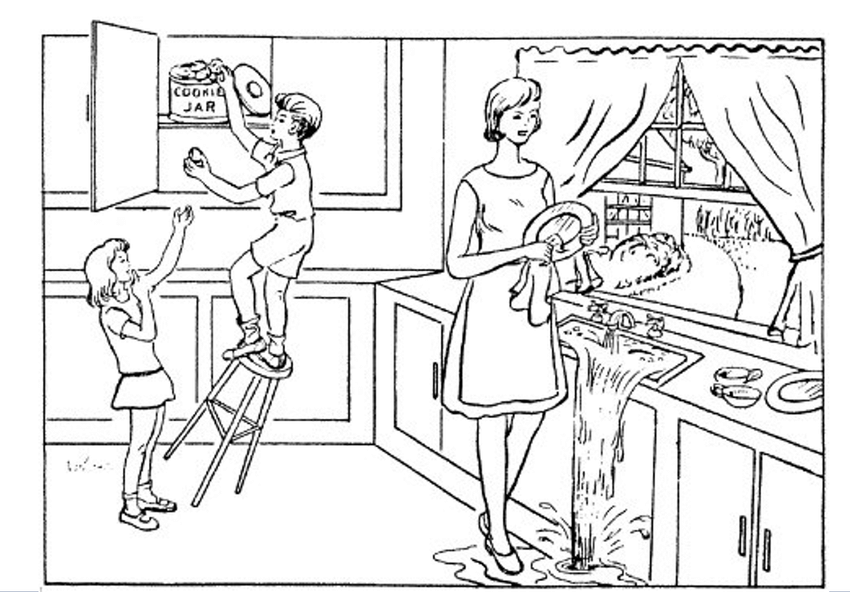}
\caption{The Cookie Theft Picture from the Boston Diagnostic Aphasia Examination."}
\label{fig:ctp}
\end{figure}

For the PD task, the examiner asks subjects to describe the picture (see Fig. \ref{fig:ctp}) by saying, "Tell me everything you see going on in this picture". Then subjects might say, "there is a mother who is drying dishes next to the sink in the kitchen. She is not paying attention and has left the tap on. As a result, water is overflowing from the sink. Meanwhile, two children are attempting to make cookies from a jar when their mother is not looking. One of the children, a boy, has climbed onto a stool to get up to the cupboard where the cookie jar is stored. The stool is rocking precariously. The other child, a girl, is standing next to the stool and has her hand outstretched ready to be given cookies.

\section{DementiaBank Pitt Corpus}
\label{app:pitt}

The dataset was gathered longitudinally between 1983 and 1988 as part of the Alzheimer Research Program at the University of Pittsburgh. The study initially enrolled 319 participants according to the following eligibility criteria: all the participants were required to be above 44 years old, have at least seven years of education, have no history of major nervous system disorders, and have an initial Mini-Mental State Examination score above 10. Finally, the cohort consisted of 282 subjects.  In particular, the cohort included 101 healthy control subjects (HC) and 181 Alzheimer’s disease subjects (AD). An extensive neuropsychological assessment was conducted on the participants, including verbal tasks and the Mini-Mental State Examination (MMSE).

\section{Architecture Overview of Models}
\label{app:architectures}
\begin{figure*}[ht]
    \centering
    \includegraphics[width=0.8\textwidth]{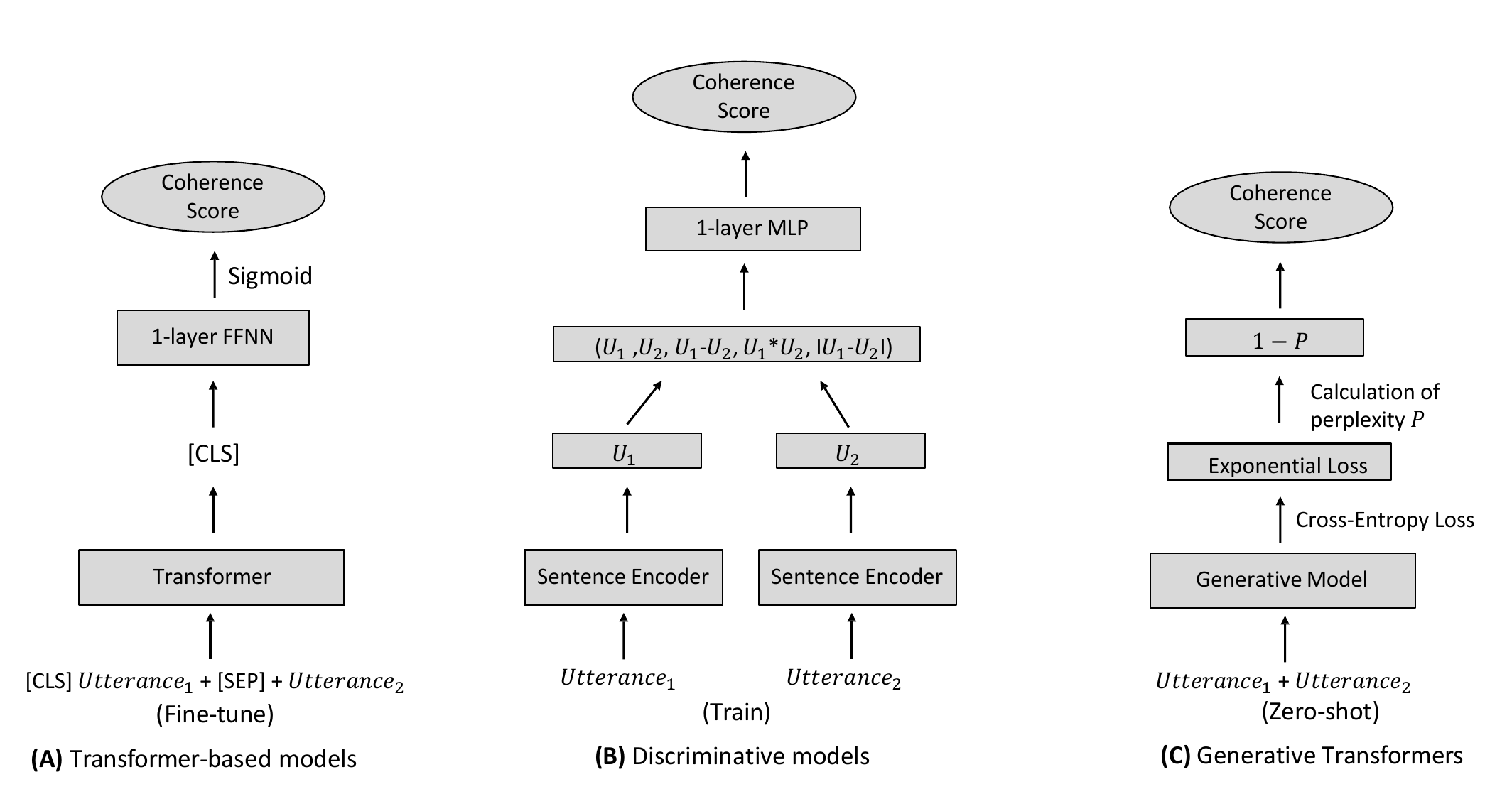}
         \caption{Architecture overview of coherence models in the three settings. The final output is always a coherence score for a given pair of sentences.}
        
    \label{fig:coherence_models}
\end{figure*}

We consider three main types of coherence models, in three different settings: a) fine-tuning transformer-based models, b) fully training discriminative models, and c) zero-shot learning with transformer-based generative models . Fig. \ref{fig:coherence_models} provides the overall architecture of coherence models in each setting. The models receive a pair of utterances in the input and output a coherence score for the given pair. The main difference between the three is that discriminative models learn constrastive patterns to obtain the probability of an utterance pair being coherent while the transformer-based models maximise the probability of the second utterance in the pair following the first.

When we experiment with zero-shot learning (Fig. \ref{fig:coherence_models} (C)), we feed each generative transformer model with adjacent pair of utterances. For calculating the probability of each word  given its preceding ones, i.e., context, we use cross-entropy loss, calculated between the genuine pair and the generated output. The exponentiation of the cross-entropy loss between the input and model predictions is equivalent to perplexity, defined as the exponentiated average negative log-likelihood of the tokenized sequence (see Eq. \ref{eq:perplexity}). A high perplexity implies a low model predictability. To this goal, we approximate the coherence as $1-P$ (see Fig. \ref{fig:coherence_models} (C)).

For CNN \cite{cui2017text}, we use pre-trained word embeddings extracted by BERT. Each pair of utterances is transformed to a 2-dimensional matrix $\in \mathcal{R}^{d \times N}$, where $d$ denotes the dimension of pre-trained BERT embeddings and $N$ is the total number of words across the pair. The rest of the architecture is similar to that one we used for transformer-based models (see Fig. \ref{fig:coherence_models} (A)). In particular, we append to the CNN module a feed-forward neural network (FFNN) followed by a sigmoid function. The coherence score is the sigmoid function of FNNN that scales the output between 0 and 1.  We trained the model by freezing the pre-trained BERT embeddings.



\section{Absolute Percentage Coherence Score Difference Formula}
\label{app:difference}
The absolute percentage difference in $f$ scores equals the absolute value of the change in $f$ between adjacent and non-adjacent sentences divided by the average of positive, i.e., $f^+$, and negative, i.e.,  $f^-$, coherence scores, all multiplied by 100, as follows: 

\begin{equation*}
   \% \Delta f = \dfrac{|\Delta f|}{ \left[ \frac{ \Sigma f}{2}  \right]} \times 100 = \dfrac{|f^+ - f^-|}{ \left[ \frac{ f^+ + f^-}{2}  \right]} \times 100
\end{equation*}

The order of the coherence scores does not matter as we are simply dividing the difference between two scores by the average of the two coherence scores.

\section{Clinical Bio-Markers}
\label{app:biomarkers}
\subsection{ Mini-Mental State Examination (MMSE)}

The Mini-Mental State Examination (MMSE) has been the most common method for diagnosing AD and other neurodegenerative diseases affecting the brain. It was devised in 1975 by Folstein et al. as a simple standardized test for evaluating the cognitive performance of subjects, and where appropriate to qualify and quantify their deficit. It is now the standard bearer for the neuropsychological evaluation of dementia, mild cognitive impairment, and AD.

The MMSE was designed to give a practical clinical assessment of change in cognitive status in geriatric patients. It covers the person’s orientation to time and place, recall ability, short-term memory, and arithmetic ability. It may be used as a screening test for cognitive loss or as a brief bedside cognitive assessment. By definition, it cannot be used to diagnose dementia, yet this has turned into its main purpose.

The MMSE includes 11 items, divided into 2 sections. The first requires verbal responses to orientation, memory, and attention questions. The second section requires reading and writing and covers ability to name, follow verbal and written commands, write a sentence, and copy a polygon. All questions are asked in a specific order and can be scored immediately by summing the points assigned to each successfully completed task; the maximum score is 30. A score of 25 or higher is classed as normal. If the score is below 24, the result is usually considered to be abnormal, indicating possible cognitive impairment. The MMSE has been found to be sensitive to the severity of dementia in patients with Alzheimer’s disease (AD). The total score is useful in documenting cognitive change over time.

\subsection{Clinical Dementia Rating (CDR)}
The Clinical Dementia Rating (CDR) is a global rating device that was first introduced in a prospective study of patients with mild “senile dementia of AD type” (SDAT) in 1982 (Hughes et al., 1982). New and revised CDR scoring rules were later introduced (Berg, 1988; Morris, 1993; Morris et al., 1997). CDR is estimated on the basis of a semistructured interview of the subject and the caregiver (informant) and on the clinical judgment of the clinician. CDR is calculated on the basis of testing six different cognitive and behavioral domains such as memory, orientation, judgment and problem solving, community affairs, home and hobbies performance, and personal care. The CDR is based on a scale of 0–3: no dementia (CDR = 0), questionable dementia (CDR = 0.5), MCI (CDR = 1), moderate cognitive impairment (CDR = 2), and severe cognitive impairment (CDR = 3). Two sets of questions are asked, one for the informant and another for the subject. The set for the informant includes questions about the subject’s memory problem, judgment and problem solving ability of the subject, community affairs of the subject, home life and hobbies of the subject, and personal questions related to the subject. The set for subject includes memory-related questions, orientation-related questions, and questions about judgment and problem-solving ability. 

\subsection{Hamilton Depression Rating (HDR)}
The Hamilton Depression Rating  (HDR) is used to quantify the severity of symptoms of depression and is one of the most widely used and accepted instruments for assessing depression. The standard version of the HDR is designed to be administered by a trained clinician, and it contains 17 items rated on either a 3- or 5-point scale, with the sum of all items making up the total score. HDR scores are classified as normal (<8), mild depression (8 to 13), mild to moderate depression (14 to 16), and moderate to severe depression (>17). The HDR may be a useful scale for cognitively impaired patients who have difficultly with self-report instruments.

\section{Training Details}
\label{app:train}

When training the coherence models, we sampled a new set of negatives (incoherent pairs of utterances) each time for a given narrative. Thus, after a few epochs, we covered the space of negative samples for even relatively long narratives. For discriminative models, we froze the sentence encoder after initialization to avoid overfitting. We run the models for 50 epochs with 4 epochs early stopping.

We used a grid search optimization technique to optimize the parameters. For consistency, we used the same experimental settings for all models. We first fine-tuned all models by performing a twenty-times grid search over their parameter pool. We empirically experimented with learning rate ($lr$): $lr \in \{0.00001,0.00002,0.00005,0.0001,0.0002\}$, batch size ($bs$): $bs \in \{16,32,64,128\}$ and optimization ($O$): $O \in \{AdamW,Adam\}$. For the discrimination models, to tune the margin hyper-parameter ($n$), we experimented with the values $n \in \{3,5,7\}$. After the fine-tuning process, we trained again all the models for 50 epochs with 4 epochs early stopping, three times. We reported the average performance on the test set for all experiments. Model checkpoints were selected based on the minimum validation loss. Experiments were conducted on two GPUs, Nvidia V-100.

\end{document}